\documentclass{article}
\usepackage{spconf,amsmath,graphicx,url}
\usepackage{xcolor,colortbl}

\definecolor{Gray}{gray}{0.85}
\newcolumntype{a}{>{\columncolor{Gray}}c}

\title{
Self-supervised learning of dense hierarchical representations for medical image segmentation
}

\name{
Eytan Kats${^\star}$
\qquad Jochen G. Hirsch ${^\dagger}$
\qquad Mattias P. Heinrich${^\star}$
}

\address{
${^\star}$ Institute of Medical Informatics, University of Lübeck, Germany\\
${^\dagger}$ Fraunhofer Institute for Digital Medicine MEVIS, Bremen, Germany 
}

\begin{document}

\maketitle

\begin{abstract}
This paper demonstrates a self-supervised framework for learning voxel-wise coarse-to-fine representations tailored for dense downstream tasks. Our approach stems from the observation that existing methods for hierarchical representation learning tend to prioritize global features over local features due to inherent architectural bias. To address this challenge, we devise a training strategy that balances the contributions of features from multiple scales, ensuring that the learned representations capture both coarse and fine-grained details. Our strategy incorporates 3-fold improvements: (1) local data augmentations, (2) a hierarchically balanced architecture, and (3) a hybrid contrastive-restorative loss function. We evaluate our method on CT and MRI data and demonstrate that our new approach particularly beneficial for fine-tuning with limited annotated data and consistently outperforms the baseline counterpart in linear evaluation settings. Our code and pre-trained models will be publicly available at \url{https://github.com/multimodallearning/hierarchical-dense-ssl}.
\end{abstract}


\section{Introduction}
\label{sec:intro}

Annotation of medical images is a time-consuming and complex task that requires expert knowledge. This hinders the development of supervised algorithms. In contrast to supervised methods, self-supervised approaches utilize unlabeled data to extract robust representations that can be fine-tuned for various downstream tasks using only a small amount of labeled data. This makes self-supervised representation learning an important area of research in the medical imaging domain.

\begin{figure}[htb]
  \centering
  \includegraphics[height=6cm]{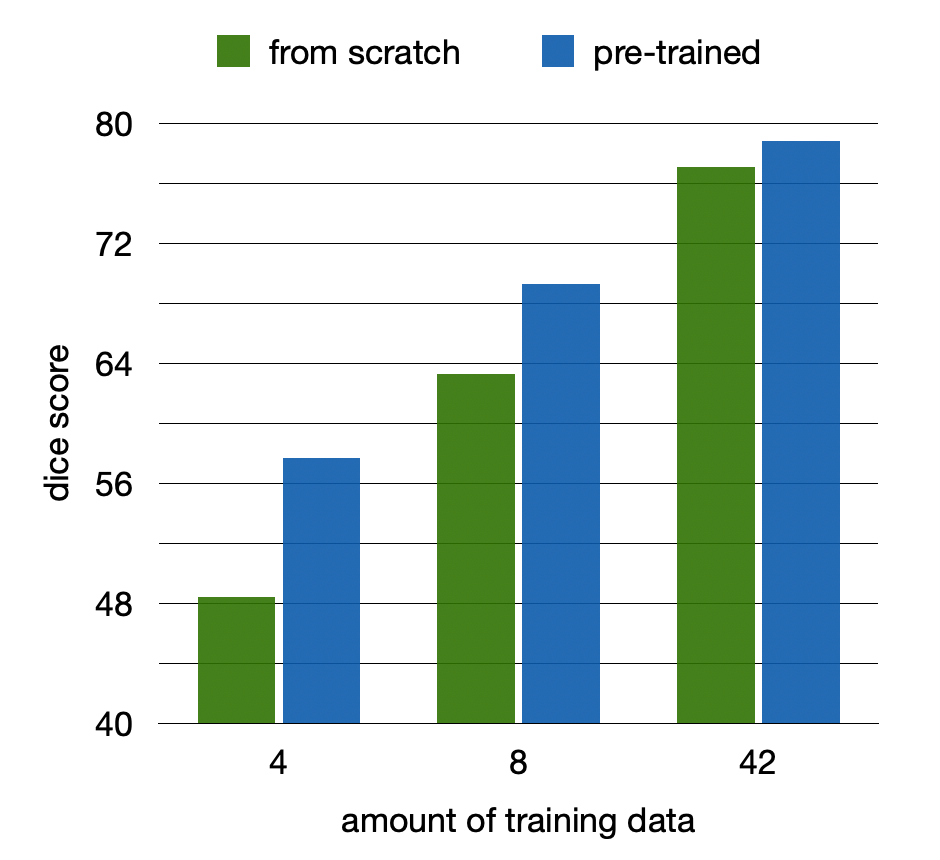}
  \caption{Fine-tuning performance for varying amounts of data. The benefit of pre-training are particularly evident when only limited labeled data is available.}
  \label{fig:fine-tuning}
\end{figure}

Contrastive and restorative approaches are two prominent techniques in self-supervised representation learning. In the restorative approach, the model learns latent features by reconstructing the original image from its augmented version \cite{He2021MaskedAA}, \cite{Chen2022MaskedIM}, \cite{Zhou2019ModelsGG}. In the contrastive approach, the model acquires robust representations by minimizing the distance between encodings of related data points (positive pairs), while pushing apart encodings of unrelated data points (negative pairs) \cite{chen2020simple}, \cite{Huang2021LesionbasedCL}, \cite{Wang2020DenseCL}. A data point can be represented by an entire image \cite{chen2020simple}, an image patch \cite{Huang2021LesionbasedCL}, or a single voxel \cite{Wang2020DenseCL}. Several methods demonstrate the effectiveness of combining contrastive and restorative approaches into a unified self-supervised framework \cite{jiang2023anatomical}, \cite{Tang2021SelfSupervisedPO}, \cite{Haghighi2022DiRADR}.

Both contrastive and restorative methods can be used to extract voxel-wise representations \cite{Zhou2019ModelsGG}, \cite{Wang2020DenseCL}, which is particularly beneficial for dense downstream tasks such as segmentation. In this context, each voxel is represented by a unique feature vector that ideally encodes all levels of information for the specific location, from fine-grained details to global context. This requires to generate feature maps in full resolution with sufficient number of channels. However, due to the memory limitations, learning dense representations that effectively capture such comprehensive information poses a challenge.

Recent studies, such as SAM \cite{Yan2020SAMSL} and vox2vec \cite{Goncharov2023vox2vecAF}, demonstrate that voxels can be effectively represented as a combination of corresponding feature vectors from different scales of a Feature Pyramid Network (FPN). SAM focuses on the landmark detection task and represents voxel using only two separate feature vectors from the highest-resolution and lowest-resolution scales. vox2vec encodes voxel as a concatenation of features from all pyramid scales and demonstrates the advantage of such representation for downstream segmentation tasks. However, the concatenated vector is characterized by a substantial imbalance between coarse and fine-grained features. The feature vector extracted from the highest-resolution scale is significantly smaller than the feature vector extracted from the lowest-resolution scale. The imbalance could result in the self-supervised training process prioritizing the optimization of low-resolution features over the high-resolution features. This may have a negative impact on the downstream segmentation task, where both capturing fine-grained details and understanding the global context are crucial for the accurate delineation.

In this paper we propose a self-supervised framework for learning hierarchical balanced coarse-to-fine representations. We adopt the vox2vec \cite{Goncharov2023vox2vecAF} approach as a baseline and introduce 3-fold improvements designed to strengthen the high-resolution component of the representation vector (Sec.\ref{sec:method}): (1) local augmentations, (2) a  hierarchically balanced architecture, and (3) a hybrid contrastive-restorative loss function. 

Our major contributions are as following: (1) We propose a self-supervised framework for learning dense hierarchical representations that effectively mitigates the intrinsic bias of the FPN-based approach towards prioritizing global context features over the fine-grained features. (2) We demonstrate that our strategy is particularly beneficial for downstream segmentation tasks with limited annotated data, and show that our pre-trained model significantly outperforms the baseline counterpart in linear evaluation setting. (3) We publicly release the pre-trained models for both CT and MRI modalities. 

\section{Method}
\label{sec:method}

The main objective of the proposed approach is to learn voxel-wise representations that are beneficial for downstream segmentation tasks. Inspired by recent advancements in coarse-to-fine modeling of dense representations \cite{Yan2020SAMSL}, \cite{Goncharov2023vox2vecAF}, we introduce a novel self-supervised training strategy (Fig.\ref{fig:framework}) that effectively addresses the inherent imbalance in FPN-based hierarchical embeddings. Local region augmentations guide the training process to learn robust fine-grained features (Sec.\ref{ssec:augmentations}). The chosen architectural configuration balances the impact of features from different scales on the contrastive loss (Sec.\ref{ssec:representations}). The reconstruction component of the hybrid contrastive-restorative loss function further emphasizes the significance of high-resolution features in the training process (Sec.\ref{ssec:loss}).

\subsection{Local augmentations}
\label{ssec:augmentations}

The effectiveness of contrastive learning is significantly dependent on the selection of suitable augmentations \cite{Zhang2022RethinkingTA}. To encourage the model to focus on fine-grained features, we incorporate local pixel shuffling and local region in-painting augmentations within a hierarchical contrastive learning framework.

First, two overlapping patches, ${X_A}$ and ${X_B}$,  are randomly cropped from a 3D image ${X}$. Then, each patch is transformed with a unique augmentation sequence ${t \sim \mathcal{T}}$ drawn from the augmentation set ${\mathcal{T}}$: ${X_A^t=t(X_A)}$, ${X_B^t=t(X_B)}$. Apart from local pixel shuffling and local region in-painting, the augmentation set ${\mathcal{T}}$ also includes non-linear intensity augmentations.  

\subsection{Balanced coarse-to-fine voxel-wise representations}
\label{ssec:representations}

\begin{figure*}[htb]
  \centering
  \includegraphics[width=18cm]{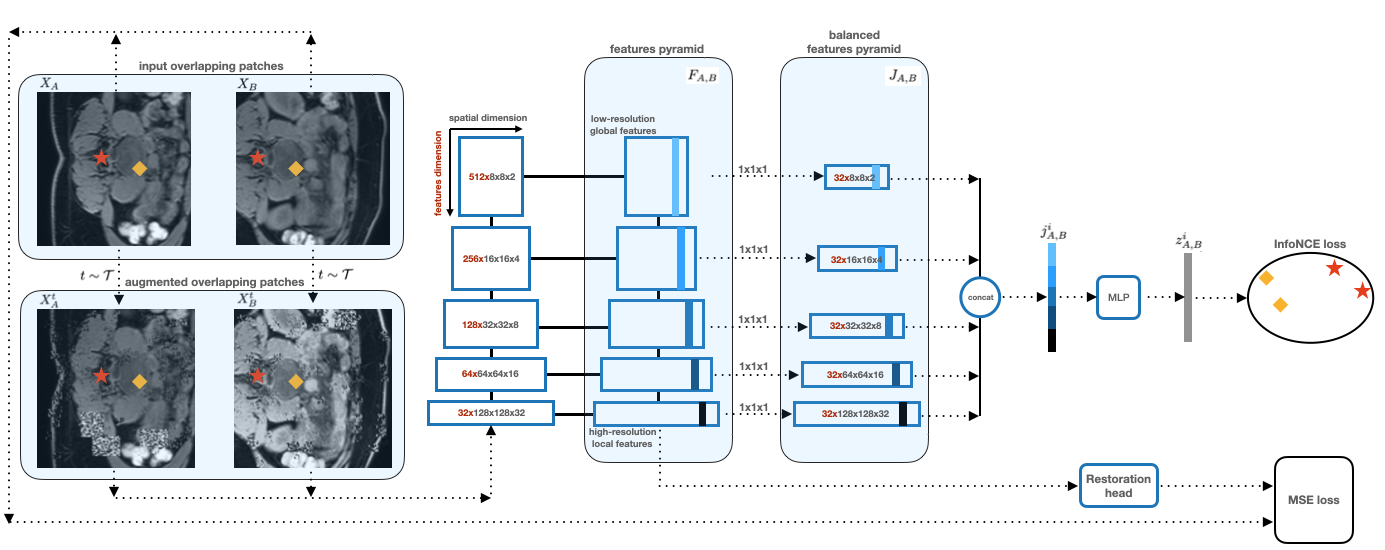}
  \caption{Pre-training pipeline. Two overlapping patches augmented and processed by the FPN. The feature maps from the different scale levels projected to the same channel size. Feature vectors sampled from the projected scale levels in corresponding positions to form hierarchically balanced voxel representation vectors. Voxel representations projected to the latent space where the InfoNCE loss is calculated.}
  \label{fig:framework}
\end{figure*}

Given two overlapping patches ${X_A^t}$ and ${X_B^t}$, the positive pair ${v_A \in X_A^t, v_B \in X_B^t}$ is defined as a pair of voxels that correspond to the same location in the original image ${X}$. Other pairs of voxels sampled from the same patch or from different patches are considered negative pairs. During each training step we randomly sample ${n}$ positive pairs from the overlapping area of patches ${X_A^t}$ and ${X_B^t}$. Each sampled voxel from the patch ${X_A^t}$ forms one positive pair with the corresponding voxel from the patch ${X_B^t}$ and ${2 \cdot (n-1)}$ negative pairs with other voxels sampled from ${X_A^t}$ and ${X_B^t}$.

FPN generates two feature pyramids that correspond to the cropped patches: ${F_A}$ and ${F_B}$ (Fig.\ref{fig:framework}). To mitigate the imbalance in voxel representation, the feature maps from different scales of each pyramid are projected to have the same number of channels resulting in balanced feature pyramids ${J_A}$ and ${J_B}$. Feature vectors are sampled from the corresponding locations of all scales in ${J_A}$ and ${J_B}$ and concatenated to create balanced voxel representations denoted as ${j_A^i}$, ${j_B^i}$, where ${i=1,..,n}$.

Comparing to the vox2vec baseline  \cite{Goncharov2023vox2vecAF} we also reduced the depth of the feature pyramid and doubled the number of feature maps in the remaining feature pyramid levels to further amplify the contribution of local features.

The proposed architecture ensures that high-resolution features have an equal contribution to the voxel representation as low-resolution features. This strategy facilitates an unbiased optimization process of both fine-grained details and global semantics, leading to enhanced performance in downstream segmentation tasks (Sec.\ref{ssec:results}). 

\subsection{Hybrid loss function}
\label{ssec:loss}

On one hand, the contrastive loss excels in learning semantically meaningful representations when applied to embeddings extracted from different levels of the feature pyramid \cite{Yan2020SAMSL}, \cite{Goncharov2023vox2vecAF}. On the other hand, the restorative approach is known to be particularly useful in capturing fine-grained details that proved to be beneficial for downstream segmentation tasks \cite{Zhou2019ModelsGG}. Therefore, we propose to include a restorative component in the hierarchical contrastive learning framework to amplify the emphasis on local, fine-grained features.

 As suggested in \cite{chen2020simple}, we first project ${j_A^i}$, ${j_B^i}$  to 128-length vectors ${h_A^i}$, ${h_B^i}$  using 3-layer fully-connected network  and normalize the result: ${z_A^i=h_A^i/||h_A^i||}$, ${h_A^i/||h_A^i||}$, ${i=1...n}$. Then we apply InfoNCE loss \cite{chen2020simple} as the contrastive objective:
\begin{equation}
L_c = -\sum_i\log\frac{d(z_A^i, z_B^i)}{d(z_A^i, z_B^i) + \sum_{j \neq i}\sum_{l\in{A,B}}d(z_k^i, z_l^j)},
\end{equation}
where $d(z_A^i, z_B^i) = \exp\left(\langle z_A^i, z_B^i \rangle/\tau\right)$, $\langle \cdot, \cdot \rangle$ is the inner product operation and $\tau$ is the temperature parameter. In our experiments $\tau=0.1$.

The feature maps of the high-resolution pyramid level processed by the reconstruction head comprising two convolution layers to produce reconstructed image $\widehat{X_A}$, $\widehat{X_B}$. We apply MSE loss as the reconstruction objective:
\begin{equation}
L_r = ((X_A - \widehat{X_A})^2 + (X_B - \widehat{X_B})^2)/2.
\end{equation}

The total loss can be formulated as:
\begin{equation}
L = L_c + \lambda L_r,
\end{equation}
where parameter ${\lambda}$ weights between contrastive and reconstruction objectives. In our experiments $\lambda=10$.

\section{Experiments}
\label{sec:experiments}

To assess the effectiveness of our approach, we conduct evaluations on both MRI and CT modalities. First, we pre-train the models on a large collection of images. Next, we use the pre-trained weights to evaluate the strength of the learned representations on downstream segmentation tasks under linear and fine-tuning settings.

\subsection{Datasets}
\label{ssec:datasets}

For the MRI pre-training, we employ 1087 image volumes covering the abdomen region from the NAKO dataset \cite{Bamberg2022WholebodyMR}. Subsequently, for MRI downstream segmentation task we utilize 60 images with annotations for 12 abdominal organs from the publicly available AMOS MRI dataset \cite{ji2022amos}. The pre-processing steps for MRI images include (1) interpolation to the voxel spacing $1.5\times1.5\times1.5 mm^3$, (2) clipping intensity values to 0.01 and 99.9 percentile, and afterwards scaling to the range between 0 and 1, (3) cropping to the minimal volume containing the voxels with the intensity greater than 0.3.

For the CT pre-training we use 2500 publicly available image volumes from the abdomen domain: 500 images from the AMOS CT dataset \cite{ji2022amos} and 2000 images from the FLARE dataset \cite{ma2022fast}. The BTCV dataset \cite{landman2015miccai} is utilised for the downstream task. It consists of 30 CT images with 13 annotated abdominal organs. Preprocessing for the CT data comprises of the following steps: (1) interpolation to the voxel spacing $1\times1\times2 mm^3$, (2)  cropping to the minimal volume containing all the voxels with the intensity greater than -500 HU, (3) clipping intensity values to -1350 HU and 1000 HU, and afterwards scaling them to the range between 0 and 1.

\subsection{Implementation details}
\label{ssec:implementation_details}

We adopt vox2vec \cite{Goncharov2023vox2vecAF} as our baseline method, as it show the state-of-the-art results for the hierarchical contrastive learning and outperforms other methods on the downstream segmentation task. This suggests its suitability as a strong baseline to demonstrate an effectiveness of ours approach. Consequently, we adopt most of the the training parameters from vox2vec.

The models are pre-trained for 50K steps using the Adam optimizer \cite{Kingma2014AdamAM} with a learning rate of $3\cdot10^{-4}$. Each training batch includes 10 pairs of overlapping patches, with 1K voxel locations sampled per pair. Training is conducted on a single A100-80GB GPU, and the process takes an average of 35 hours. For downstream tasks we train models for 45K steps with batch size 7 employing the Adam optimizer with a learning rate of $3\cdot10^{-4}$.

In the fine-tuning configuration the non-linear head is attached to the FPN backbone. The backbone is initially frozen for 15K steps, after which the learning rate is exponentially increased for the backbone parameters from $3\cdot10^{-5}$ to $3\cdot10^{-4}$ during 1,200 steps.

In the linear evaluation setup we attach the linear head to the FPN backbone. Each pyramid level is separately fed to a dedicated convolutional layer with kernel size 1 which maps the number of channels to the number of classes. Subsequently, all the outputs are up-sampled to the full resolution and summed up.

\subsection{Results}
\label{ssec:results}

For the experiments with the AMOS MRI dataset we employ a 5-fold cross validation, and a 3 fold cross-validation for the experiments with the BTCV CT dataset. Performance of all models is evaluated using the Dice score. For all experiments we report mean results of cross-validation for the abdominal organs: spleen $Sp$, kidneys $Kid$, gallbladder $Gb$, esophagus $Es$, liver $Li$, stomach $St$, aorta $Aor$, inferior vena cava $IVC$, pancreas $Pa$ and adrenal glands $AG$. We also report the mean and standard deviation of the overall Dice score across folds.
\begin{table}[htb]
    \centering
    \resizebox{8.5cm}{!}{
    \begin{tabular}{c c c c|c|c|c|c|c|c|c|c|c|c|c}
     aug & $L_c$ & $L_r$ & arch & Sp & Kid & Gb & Es & Li & St & Aor & IVC & Pa & AG & Overall\\ \hline
     & $\surd$ & & & 76.9 & 79.3 & 33.2 & 35.1 & 87.8 & 51.1 & 73.0 & 62.0 & 55.3 & 36.4 & 58.8 $\pm$ 3.5\\
     $\surd$ & $\surd$ & & & 78.9 & 82.7 & 38.9 & 40.8 & 89.1 & 50.9 & 77.9 & 64.6 & 59.1 & 39.7 & 62.1 $\pm$ 2.4\\
     $\surd$ & & $\surd$ & & 57.3 & 62.3 & 30.7 & 25.6 & 80.0 & 33.6 & 59.2 & 44.9 & 36.1 & 20.5 & 44.4 $\pm$ 4.3\\
     $\surd$ & $\surd$ & $\surd$ & & 79.7 & 82.7 & 40.0 & 41.2 & 90.2 & \textbf{56.6} & 78.4 & 63.0 & \textbf{62.8} & 39.0 & 62.9 $\pm$ 2.6\\
     $\surd$ & $\surd$ & $\surd$ & $\surd$ & \textbf{83.2} & \textbf{84.9} & \textbf{44.9} & \textbf{47.4} & \textbf{90.3} & 56.1 & \textbf{80.6} & \textbf{69.2} & 62.0 & \textbf{42.5} & \textbf{65.7 $\pm$ 2.6}\\
    \end{tabular}}
    \caption{Ablation study of the proposed method in linear evaluation settings with MRI data.}
    \label{tab:ablation_study}
\end{table}

First, to assess the impact of individual components on the final performance of our method we conduct ablation experiments, employing linear evaluation settings and MRI data (Table \ref{tab:ablation_study}). We start from the baseline vox2vec \cite{Goncharov2023vox2vecAF} model that contains only contrastive loss function $L_c$ and evaluate different combinations of the proposed modifications: local region augmentations $aug$, restorative loss function $L_r$ and architectural changes $arch$. Remarkably, addition of each of these individual components consistently lead to performance improvement. The results confirm the importance of each modification in mitigating the inherent imbalance of hierarchical contrastive learning.

Next we evaluate our method in linear evaluation settings for CT data (Table \ref{tab:linear_evaluation}). Our method outperforms the state-of-the-art baseline approach \cite{Goncharov2023vox2vecAF} for models pre-trained on equal amounts of data. Moreover, our model surpasses a model initialized with the publicly released vox2vec weights, which was trained on 2.5 times more data.

\begin{table}[htb]
    \centering
    \resizebox{8.5cm}{!}{
    \begin{tabular}{c|c|c|c|c|c|c|c|c|c|c|c|c|c}
     method & data & Sp & Kid & Gb & Es & Li & St & Aor & IVC & PVC & Pa & AG & Overall\\ \hline
     vox2vec \cite{Goncharov2023vox2vecAF} & 6550 & 80.9 & 81.0 & 30.6 & 58.3 & \textbf{90.4} & \textbf{62.3} & 83.3 & 74.0 & 52.8 & 48.1 & 47.2 & 64.4 $\pm$ 1.3\\
     vox2vec \cite{Goncharov2023vox2vecAF} & 2500 & 79.3 & 79.0 & 24.2 & 53.4 & 88.6 & 57.9 & 82.6 & 72.7 & 50.5 & 46.8 & 46.0 & 62.0 $\pm$ 1.0\\
     ours & 2500 & \textbf{81.4} & \textbf{83.1} & \textbf{32.5} & \textbf{58.4} & 89.5 & 59.5 & \textbf{83.6} & \textbf{75.6} & \textbf{54.8} & \textbf{54.5} & \textbf{48.4} & \textbf{65.6 $\pm$ 0.8}\\
    \end{tabular}}
    \caption{Results of linear evaluation on CT data. We employed 2500 images to pre-train model with our method and vox2vec method, and compared their performance with publicly available vox2vec model which was pre-trained using 6550 images.}
    \label{tab:linear_evaluation}
\end{table}

Finally, we conduct a comparison between model fine-tuning and training from scratch with varying amounts of annotated data (Table \ref{tab:finetuning_vs_fromscratch}). We observe that fine-tuning consistently outperforms training from scratch, while especially beneficial in low-data scenarios. The result highlights the strength of learned hierarchical representations, which can be effectively adapted to specific tasks even with limited training data. This property is particularly important in the medical domain where annotated data remains scarce.

\begin{table}[htb]
    \centering
    \resizebox{8.5cm}{!}{
    \begin{tabular}{c|c|c|c|c|c|c|c|c|c|c|c}
     method & Sp & Kid & Gb & Es & Li & St & Aor & IVC & Pa & AG & Overall\\ \hline
     \multicolumn{12}{c}{42 training images}\\ \hline
     from scratch & 89.3 & 90.7 & 59.6 & 61.0 & 93.7 & 81.1 & 89.5 & 84.4 & 76.9 & 54.4 & 77.1 $\pm$ 5.4\\
     fine-tuning & \textbf{90.8} & \textbf{91.7} & \textbf{67.0} & \textbf{62.3} & \textbf{94.5} & \textbf{81.9} & \textbf{89.6} & \textbf{84.9} & \textbf{79.1} & \textbf{56.2} & \textbf{78.8 $\pm$ 4.3}\\ \hline
     \multicolumn{12}{c}{8 training images}\\ \hline
     from scratch & 81.3 & 80.6 & 34.5 & 47.0 & 90.7 & 65.7 & 80.8 & 69.3 & 57.5 & 35.6 & 63.3 $\pm$ 6.7\\
     fine-tuning & 86.8 & 85.6 & 50.8 & 46.0 & 93.0 & 66.9 & 84.6 & 76.7 & 67.8 & 44.3 & 69.3 $\pm$ 4.7\\ \hline
     \multicolumn{12}{c}{4 training images}\\ \hline
     from scratch & 66.2 & 67.7 & 18.7 & 35.0 & 85.1 & 40.7 & 72.1 & 54.9 & 39.2 & 16.6 & 48.4 $\pm$ 10.1\\
     fine-tuning & 81.2 & 78.6 & 20.8 & 41.0 & 90.9 & 52.7 & 78.3 & 63.6 & 51.1 & 27.5 & 57.7 $\pm$ 7.8\\
    \end{tabular}}
    \caption{Comparison of fine-tuning with training from scratch on MRI data of various sizes: 42, 8, and 4 training images.}
    \label{tab:finetuning_vs_fromscratch}
\end{table}

\section{Conclusions}
\label{sec:conclusions}

We propose a self-supervised framework for learning hierarchically balanced voxel-wise representations from unlabeled data. Our approach effectively mitigates the inherent imbalance of FPN-based embeddings, ensuring that both high-resolution and low-resolution features contribute equally to the learned voxel representations. We demonstrate that our method outperforms the baseline in linear evaluation settings and demonstrate that it is particularly advantageous in fine-tuning settings with limited labeled data.

\section{Compliance with ethical standards}
\label{sec:ethics}

The German National Cohort (NAKO) study is performed with the approval of the relevant ethics committees, and is in accordance with national law and with the Declaration of Helsinki of 1975 (in the current, revised version).

\section{Acknowledgments}
\label{sec:acknowledgments}

We gratefully acknowledge the financial support by German Research Foundation: DFG, HE 7364/10-1, project number 500498869.
 
\bibliographystyle{IEEEbib}
\bibliography{ISBI2024_HBSSL}

\end{document}